\documentclass[10pt,twocolumn,letterpaper]{article}

\usepackage{cvpr}
\usepackage{times}
\usepackage{epsfig}
\usepackage{amssymb}
\usepackage{framed,multirow}

\usepackage{latexsym}
\usepackage{amsmath,amsfonts}
\usepackage{graphicx}
\usepackage{setspace}
\usepackage{tocloft}
\usepackage{subcaption}
\usepackage{float}
\usepackage{multicol}
\usepackage{booktabs}
\usepackage{wrapfig}
\newcommand{\norm}[1]{\left\lVert#1\right\rVert}
\usepackage{url}
\usepackage{xcolor}
\definecolor{newcolor}{rgb}{.8,.349,.1}
\usepackage[pagebackref=true,breaklinks=true,letterpaper=true,colorlinks,bookmarks=false]{hyperref}

\cvprfinalcopy % *** Uncomment this line for the final submission

\def\cvprPaperID{****} % *** Enter the CVPR Paper ID here

\ifcvprfinal\fi
\begin{document}

%%%%%%%%% TITLE
\title{LABNet: Local Graph Aggregation Network with Class Balanced Loss for Vehicle Re-Identification}

% \author{Abu Md Niamul Taufique \and Andreas Savakis}
% Rochester Institute of Technology \\ Rochester, NY, 14623
% \author{Andreas Savakis}
% \and Andreas Savakis \\
% Rochester Institute of Technology\\
% Rochester, NY, USA, 14623\\
% {\tt\small at7133@rit.edu, andreas.savakis@rit.edu}
% }
\author{Abu Md Niamul Taufique \quad Andreas Savakis\\Rochester Institute of Technology \\ Rochester, NY, 14623\\ at7133,andreas.savakis@rit.edu}

% \and
% \author{Abu Md Niamul~Taufique,
%         Andreas~Savakis,~\IEEEmembership{Senior Member,~IEEE}}
\maketitle

\begin{abstract}
Vehicle re-identification is an important computer vision task where the objective is to identify a specific vehicle among a set of vehicles seen at various viewpoints. Recent methods based on deep learning utilize a global average pooling layer after the backbone feature extractor, however, this ignores any spatial reasoning on the feature map. In this paper, we propose local graph aggregation on the backbone feature map, to learn associations of local information and hence improve feature learning as well as reduce the effects of partial occlusion and background clutter. Our local graph aggregation network considers spatial regions of the feature map as nodes and builds a local neighborhood graph that performs local feature aggregation before the global average pooling layer. We further utilize a batch normalization layer to improve the system effectiveness. Additionally, we introduce a class balanced loss to compensate for the imbalance in the sample distributions found in the most widely used vehicle re-identification datasets. Finally, we evaluate our method in three popular benchmarks and show that our approach outperforms many state-of-the-art methods. 
\end{abstract}

\section{Introduction}
Vehicle Re-IDentification (VRID) has significant applications in smart traffic control systems, video surveillance and visual target tracking \cite{liu2016deep, lou2019veri, taufique2020benchmarking}. 
The objective of VRID is to find a vehicle of interest (probe) within a set of vehicles (gallery set) in a single camera or multi camera settings. Various real world scenarios make the VRID problem challenging such as occlusion, minor appearance differences among different vehicles, illumination variations, pose variations, etc. 

\begin{figure}[t!]
\centerline{\includegraphics[width=0.3\textwidth]{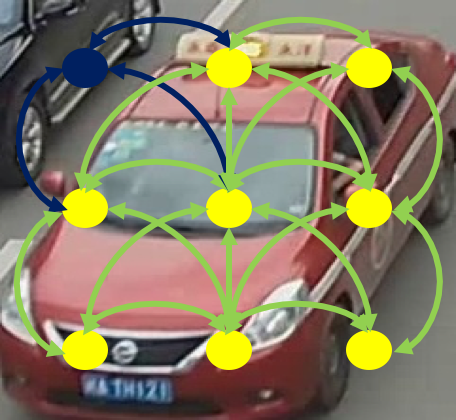}}
\caption{Illustration of spatial information aggregation used by graph network for smoothing spatial features.}
\label{fig:concept}
\end{figure}

In recent years significant progress has been made in VRID  with the advancement of deep learning techniques \cite{he2019part, meng2020parsing, zheng2020vehiclenet, khorramshahi2020devil} and the introduction of large VRID benchmarks \cite{lou2019veri, liu2017provid, liu2016deep}. In all of these benchmarks, images are captured in a multi-camera settings to get multiple instances of the same vehicle under various pose and illumination conditions. However, in several scenarios, a vehicle might not appear in multiple cameras at the same time or some vehicles are captured more frequently than others. As a result, these benchmarks contain
% various vehicles with an 
imbalanced sample distributions. 
In this work, we propose a local graph aggregation network module that associates spatial features (see Fig. \ref{fig:concept}) and incorporate  a class balanced loss to overcome the class imbalance in the training dataset. 

Zheng et al \cite{zheng2020vehiclenet} showed that the limited number of samples present in several ids may hinder the learning procedure and proposed to use multiple datasets simultaneously to learn a more generalized feature representation. Zheng et al \cite{zheng2020going} proposed to utilize generative techniques to generate new samples with different viewpoints. The additional training images help achieve pose invariant features and hence better performance. However, both of the aforementioned approaches require additional training data that may be difficult or time consuming to acquire.

Several methods have been proposed to utilize detection and matching of license plate information for VRID \cite{anagnostopoulos2008license, watcharapinchai2017approximate} as the license plate number is unique for each vehicle. However, license plates may not be visible in various circumstances based on the pose of the vehicle or partial occlusion. This suggest that learning visual cues from the vehicle appearance would be key to VRID. 
The initial works for appearance-based VRID involved hand crafted feature extraction techniques to extract pose and illumination invariant features \cite{liu2016large, zapletal2016vehicle, khan2019survey}. 
Recently, deep learning techniques gained significant popularity due to their superior performance \cite{he2019part, meng2020parsing, zhu2019vehicle, khorramshahi2019dual}.   

% However, the challenge appears when the objective is 
Despite recent progress, it remains a challenge to learn visual cues that are invariant under intra-class pose, illumination and partial occlusion variations, but also discriminative to the subtle inter-class differences in appearance \cite{meng2020parsing}. 
Chu et al \cite{chu2019vehicle} proposed to tackle the problem of intra-class pose variation using a viewpoint-aware metric learning approach, where they learn two feature spaces for similar and different viewpoints. 
Meng et al \cite{meng2020parsing} proposed to utilize a network for parsing vehicles into multiple views and then improvise the feature representation based on the parsed mask to align the features for four different viewpoints. 
Tang et al \cite{tang2019pamtri} proposed another method that utilizes synthetic data to estimate vehicle pose and shape information to learn pose invariant features by learning the vehicle type and color information in a multitask learning framework. 
Chen et al \cite{chen2020orientation} proposed orientation invariant feature learning using semantic guided part attention. 
He et al \cite{he2019part} proposed to use part detection and attention mechanism to learn fine grained feature.
However, all of these methods require extracting additional elements, such as part detection or segmentation, to improve pose invariant properties. 
Chen et al \cite{chen2019partition} proposed to split the output feature map of the backbone and learn the spatial significance of the feature map explicitly without any additional annotation of specific parts of the vehicle or keypoints. 
To achieve this, they partitioned the backbone feature map into multiple partitions and computed losses for these multiple branches. This partition strategy showed improved VRID performance. However, it may suffer due to  part missalignment for various poses of the same vehicle \cite{shen2020exploring}.

% \textcolor{red}{}
In this research, we proposed a Local graph Aggregation Network with class Balanced loss (LABNet) for VRID. 
We introduce a local graph aggregation module to learn spatial relationships in the feature map. Local graph aggregation improves pose invariant feature learning under partial occlusion or in complex background scenarios. 
% In our framework, we construct a spatial graph on the extracted backbone feature map and build a local graph with it. 
A simple illustration of our framework is shown in Fig. \ref{fig:concept}, where a spatial graph is constructed on the extracted backbone feature map (actual constellation is $20\times20)$. 
Our results demonstrate that stacking graph aggregation modules improves learning under partial occlusion, background clutter and pose variations.
% the aggregation of these graph nodes helps improves learning over intra-class pose variation as well as overcome challenges such as partial occlusion or complex background. 
We also introduce a class balanced loss along with triplet id loss during network training. The objective is to improve the generalization capabilities of the network by putting more weight on classes with fewer samples. 
% Most of the recent 
Current methods overlook the class imbalance problem during training, which is known to bias the system towards classes with the largest number of samples. 
The main contributions of our work are outlined below.
\begin{enumerate}
    \item We propose a novel parameter-free Local Graph Aggregation (LGA) module after feature extraction for information diffusion among spatial nodes.
    \item We present the LABNet architecture for VRID that includes a cascade of LGA modules to increase the receptive field of the feature map and a batch normalization layer for more effective training.
    \item We introduce a class balanced loss, that is used in combination with triplet loss for VRID, to tackle the class imbalance problem in the training dataset.
    \item Our method is extensively tested and outperforms state-of-the-art methods on several datasets.
\end{enumerate}

The rest of the paper is organized as follows. 
In Section \ref{sec:related_work}, we discuss related works. 
In section \ref{sec:method}, we present the details of our method. 
In section \ref{sec:datasets}, we discuss various VRID benchmarks used for testing and present the experimental details.
In section \ref{sec:evaluation_metrics}, we discuss the evaluation metrics we use to quantify VRID performance. 
In section \ref{results}, we show qualitative results and compare our method to state-of-the-art methods. 
Finally, in section \ref{sec:conclusion}, we present final remarks.

\section{Related Work}
\label{sec:related_work}
% % Several methods have been published in the literature for VRID. Some of the existing methods related to our method are discussed below.
% % \subsection{Augmentations}
% % Collecting and annotating VRID datasets is a time consuming and expensive task.
% % On the other hand, learning
% Learning pose invariant features requires large labeled datasets, as training deep networks on a small datasets
% % with deep learning techniques
% often leads to overfitting. 
% However, collecting and annotating large VRID datasets is a time consuming and expensive task.
% To address these issues, various augmentation techniques are used during training deep learning models for VRID. 
% Popular augmentation techniques such as random cropping, random flipping, random rotation, random scaling are often used for training VRID models. Also, in VRID task, random erasing \cite{shen2020exploring, khorramshahi2020devil, he2020multi} augmentation technique shows significant performance gain.
% \textcolor{red}{}
% \subsection{Vehicle Re-identification}
% Recently, 
Several VRID methods that utilize deep learning techniques have been proposed in recent years.
Bai et al \cite{bai2018group} proposed a group-sensitive triplet embedding technique to efficiently train deep learning algorithms where the intra-class variance is modeled with an intermediate representation group. 
% For the same vehicle ids, the distances among these groups are minimized where the distances are maximized for the groups with different vehicle ids. 
Kuma et al \cite{kuma2019vehicle} compared various sampling techniques during the triplet loss calculation and showed that utilizing proper sampling technique for triplet loss along with id loss can significantly improve VRID performance. 
He et al \cite{he2020multi} proposed to use a multi-domain learning and identity mining to improve the VRID performance. This approach adopted a baseline method \cite{luo2019strongtmm} that uses soft margin triplet loss and id loss during training. 
P. Khorramshahi et al \cite{khorramshahi2020devil} used a self-supervised residual generation technique as an attention mechanism to perform VRID.
Porrello et al \cite{porrello2020robust} proposed a teacher-student framework and multi-view knowledge distillation to tackle the viewpoint variation in VRID.
% Network diagram
\begin{figure*}[htbp]
\centerline{\includegraphics[width=0.9\textwidth]{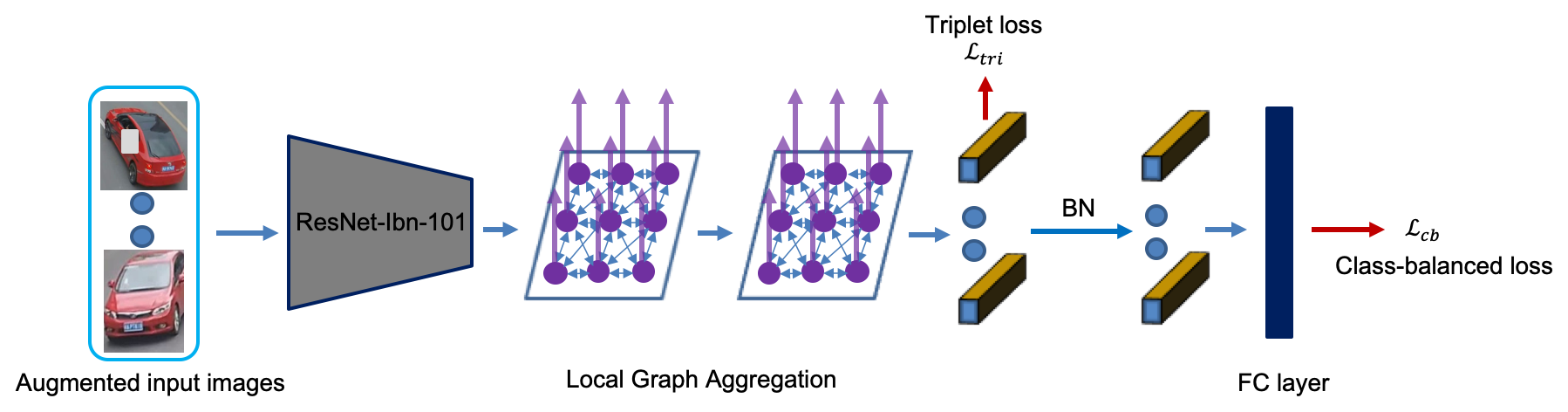}}
\caption{Proposed network architecture. Best viewed in color.}
\label{fig:network}
\end{figure*}

\subsection{Feature aggregation using graph networks}
Recently, various research efforts utilize graph based reasoning for VRID and person re-identification (PRID). 
The Pyramidal Graph Network (PGN) \cite{shen2020exploring} is a multiscale pyramidal graph network to incorporate multiscale information from backbone features. 
% The research shows that Utilizing spatial graph networks is beneficial for VRID. 
The multiscale pyramid is built utilizing various spatial pooling layers and using a Graph Convolution Network (GCN) to diffuse information within the pooled feature maps. 
% Utilizing spatial graph networks is found beneficial for VRID. 
Finally, the feature maps from these stages are summed and passed to the classifier to compute the classification score. The overall graph-based information aggregation shows improved performance over the baseline.  
The Masked Graph Attention Network (MGAT) \cite{bao2019masked} utilized graph based information processing on a minibatch of images, where features of each image are considered as a node and their mutual information is processed to learn an updated representation for PRID.

Another approach for aggregating features is to use part-level vehicle information. For part-level information aggregation, horizontal or vertical parts of the backbone feature map are aggregated and the loss is computed for all the part-level features \cite{quan2019auto}. This type of feature aggregation helps with pose invariant and fine grained feature learning for VRID. 
In this work, we consider local neighborhood graph aggregation for explicit spatial reasoning. We also utilize multiple graph modules that are stacked to accumulate information within a larger local region.
% radius defined during graph edge formulation. 

\subsection{Loss functions}
In various VRID tasks, adding multiple loss functions generally helps to achieve improved performance. The most commonly used loss function is the id loss, or softmax loss function \cite{sun2014deep},
% . Id loss is 
that achieves better convergence.
Label smoothing is another component that is mostly used in combination with the id loss to alleviate overfitting \cite{khorramshahi2020devil, shen2020exploring} in VRID tasks.
Another popular loss function for VRID is the triplet loss \cite{schroff2015facenet}. Several works have utilized both the triplet loss and the id loss to train the VRID algorithms \cite{khorramshahi2020devil,shen2020exploring,hermans2017defense}. However, existing VRID methods do not address the dataset imbalance problem with a class balances loss, such as the one introduced in \cite{cui2019class}.

\subsection{Dataset augmentation and balance}
% Several methods have been published in the literature for VRID. Some of the existing methods related to our method are discussed below.
% \subsection{Augmentations}
% Collecting and annotating VRID datasets is a time consuming and expensive task.
% On the other hand, learning
Learning pose invariant features requires large labeled datasets, as training deep networks on a small datasets
% with deep learning techniques
often leads to overfitting. 
However, collecting and annotating large VRID datasets is a time consuming and expensive task.
To address these issues, various augmentation techniques are used during training deep learning models for VRID. 
Popular augmentation techniques such as random cropping, random flipping, rotation and scaling are often used for training VRID models. Additionally, random erasing augmentation \cite{shen2020exploring, khorramshahi2020devil, he2020multi} shows significant performance gain. 

Furthermore, dataset imbalance is a significant issue with VRID , as illustrated by the distribution of ids shown in Fig. \ref{fig:data_dist}. 
The problem of class imbalance is widely studied in machine learning 
\cite{johnson2019survey},
as it often leads to classification bias.
We adapt the class balanced loss proposed in \cite{cui2019class} that is designed to counteract the effects of dataset imbalance.

\section{Methodology}
\label{sec:method}
The overall architecture of our method is shown in Fig. \ref{fig:network}. The input images are passed through the backbone feature extractor and the local graph is built on top of the backbone feature map. Then the aggregated features are used to compute the triplet loss function. A batch normalization (BN) layer is included to improve the effectiveness of training and it is followed by a fully connected (FC) layer. The output of the FC layer is used to compute the class balanced loss.

\subsection{Local graph based feature aggregation}
 We consider an input sample $X \in \mathbb{R}^{3\times W \times H}$, where $W$ and $H$ are the image width and height respectively, that is processed through the backbone. 
 The output feature map can be written as $\mathbf{x} \in \mathbb{R}^{c \times w \times h}$ where $c$ is the feature dimensions, $w$ is the width and $h$ is the height of the feature map. The width and height of the feature map depend on the pooling, padding, and stride of the backbone network. 
 We build a graph $G(\mathcal{V},\mathcal{E})$ with nodes $\mathcal{V} \in \mathbb{R}^{k}$ and edges $\mathcal{E} \in \mathbb{R}^{k \times k}$, where $k=w \times h$. 
 
For the edge formation, we assume the output backbone feature map, of width and height $w \times h$, lies in a Euclidean coordinate space as depicted in Fig. \ref{fig:network}.  
For node $i \in \mathcal{V}$ with all possible edges $(j,i) \in \mathcal{E}$ where $j \in \mathcal{V}$, the neighborhood set can be written as follows.
\begin{equation}
    \mathcal{N}(i) = \{j | d(p_i,p_j) < r \}, \forall j, j \neq i
\end{equation}
where, $p_i$ and $p_j$ are coordinate locations of node $i$ and $j$ respectively,
$d(.,.)$ denotes Euclidean distance, and $r$ is the distance threshold for the edge formation.
% We consider the origin at the top left as shown in Figure \ref{fig:network}.
The $n^{th}$ aggregation layer \cite{FeyLenssen2019} can be written as follows.
\begin{equation}
    \mathbf{x}_i^n = 
    ReLU\left(\sum_{j \in \mathcal{N}(i)\cup \{i\}} \frac{1}{\sqrt{deg(i)} \cdot \sqrt{deg(j)}} \cdot \left( \mathbf{x}_j^{(n-1)}\right)\right)
\end{equation}

% In a simplistic form, if 
We consider the graph $G(\mathcal{V}, \mathbf{A})$, where $\mathbf{A} \in \mathbb{R}^{k \times k}$ is an affinity matrix, ane the aggregation function can be written as follows.
\begin{equation}
    \mathbf{x}_a = ReLU(\Tilde{\mathbf{D}}^{-\frac{1}{2}}\Tilde{\mathbf{A}}\Tilde{\mathbf{D}}\mathbf{x})
\end{equation}
\noindent
where $\Tilde{\mathbf{A}}=\mathbf{A}+\mathbf{I}$ is the affinity matrix with the self loops, $\Tilde{\mathbf{D}} \in \mathbb{R}^{k\times k}$ is the degree matrix of $\Tilde{\mathbf{A}}$ and $\mathbf{x}_a$ is the output aggregated feature and $ReLU$ is the non-linearity operation. 
The propagation rule is adapted from \cite{kipf2016semi}.

% Increasing $r$ to increase 
The receptive field of the aggregated feature map can be increased by increasing $r$, but that adds significant computational overhead and requires large memory. 
To overcome this limitation, we add multiple LGA layers in cascade that efficiently increase the receptive field of the feature map. 
With this arrangement, we can effectively increase the receptive field of the aggregated feature map without adding significant computational overhead. 
The output of the last LGA layer is propagated through a global average pooling (GAP) layer and directly used for the triplet loss computation. 
We implemented several types of graph-based feature aggregation techniques and found that parameter-free aggregation yields the best results. 
We found out that adding a weight matrix during the graph propagation caused overfitting of the model and hurt the final VRID performance. 
We also considered various graph pooling techniques to combine the outputs of nodes from the backbone feature map but this did not improve performance.

% \subsection{Batch normalization layer before classifier}
Similar to previous studies \cite{khorramshahi2020devil, he2020multi}, we  found that utilizing a batch normalization (BN) layer before the FC layer yields the best performance. 
We include a BN layer after the final LGA module and before the FC layer as depicted in Fig. \ref{fig:network}.

\begin{figure*}[hbtp]
\centerline{\includegraphics[width=0.99\textwidth]{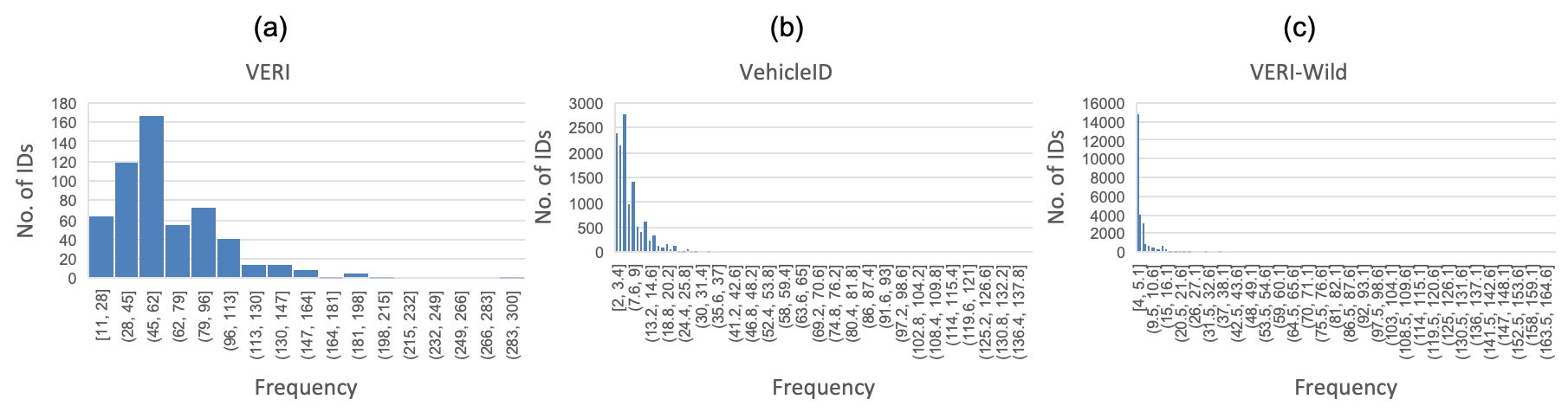}}
\caption{Imbalance in sample distributions for (a) VERI, (b) VehicleID, (c) VERI-Wild benchmarks. Best viewed in color.}
\label{fig:data_dist}
\end{figure*}

\begin{table*}[htbp]
\caption{Dataset description} 
\label{tab:dataset}
\centering
{\begin{tabular}{cc|c|ccc|ccc} 
\hline
\hline
\multicolumn{2}{c|}{\textbf{Dataset}}& \textbf{VERI} & \multicolumn{3}{c|}{\textbf{VehicleID}} & \multicolumn{3}{c}{\textbf{VERI-Wild}} \\
\hline
\multirow{2}{*}{Train} & Images & 37,746 & \multicolumn{3}{c|}{110,178} & \multicolumn{3}{c}{277,797} \\
& Vehicles & 576 & \multicolumn{3}{c|}{13,164} & \multicolumn{3}{c}{30,671}  \\
\hline
\multirow{4}{*}{Test} & Subset & x & small & medium & large & small & medium & large \\
& Gallery & 11,579 & 800 & 1,600 & 2,400 & 41,816 & 69,389 & 138,517 \\
& Probe & 1,678 & 5,693 & 11,777 & 17,377 & 3,000 & 5,000 & 10,000 \\
& Vehicles & 200 & 800 & 1,600 & 2,400 & 3,000 & 5,000 & 10,000 \\
\hline
\multicolumn{2}{c|}{Camera} & 20 & \multicolumn{3}{c|}{12} & \multicolumn{3}{c}{174} \\
\hline
\multicolumn{2}{c|}{Views} & 6 & \multicolumn{3}{c|}{2} & \multicolumn{3}{c}{unconstrained} \\
\hline
\hline
\end{tabular}}
%\end{center}
\end{table*}

\subsection{Loss functions}
The output of the local graph module is used for the computation of the triplet loss \cite{he2020multi} that is constructed as follows.
\begin{equation}
    \mathcal{L}_{tri} = \log \left[ 1 + exp\left( \norm{\mathbf{x}^a - \mathbf{x}^p}_2^2 - \norm{\mathbf{x}^a - \mathbf{x}^n}_2^2 + m\right)\right]
\end{equation}

To tackle the class imbalance problem for id loss, we propose to adapt the class balancing loss proposed in \cite{cui2019class}. For an input sample $\mathbf{X}$ the output score of the network with one graph module can be written as follows.
\begin{equation}
    \mathbf{z} = FC(BN(G(BF(\mathbf{X}))))
\end{equation}
where, $BF$ is the backbone feature extraction network, $G$ is the local graph aggregation module, BN is the batch normalization layer and $FC$ is the fully connected layer.

% Additionally, considering the ground truth label is $y \in {1,2,.....,T}$ and $T$ is the number of classes,
The class balanced id loss  \cite{cui2019class} can be written as follows.
% \textcolor{red}{}
\begin{equation}
\label{eq:CBloss}
    \mathcal{L}_{CB}(\mathbf{z}, y) = - \frac{1-\beta}{1-\beta^{n_{y}}} \log \frac{\exp(\mathbf{z}_{y})}{\sum_{t=1}^{T}\exp(\mathbf{z}_t)}
\end{equation}
\noindent
where $T$ is the number of classes and the ground truth label is $y \in {1,2,.....,T}$.
The overall loss function of LABNet is the summation of the class balanced loss and the triplet loss that is given as follows. 
\begin{equation}
    \mathcal{L} = \mathcal{L}_{CB}(\mathbf{z},y) + \mathcal{L}_{tri}
\end{equation}

% \begin{equation}
%     CB(\mathbf{z}, y) = - \frac{1-\beta}{1-\beta^{n_{y}}} \log \frac{\exp(\mathbf{z}_{y})}{\sum_{j=1}^{T}\exp(\mathbf{z}_j)}
% \end{equation}

In Eq. (\ref{eq:CBloss}), the term multiplying the cross-entropy loss is a smoothing term for class balancing. The hyperparameter $\beta \in [0,1)$ and $n_y$ is the class frequency in the entire training set. One interesting property of the weighting term is that with $\beta =0$, there is no re-weighting and with $\beta \to 1$, the re-weighting is the inverse of class frequency. 
This property allows the flexibility to tune the $\beta$ parameter and optimize the training procedure for  given dataset.
% \textcolor{red}{
% The overall loss function of LABNet is the summation of the class balanced loss and the triplet loss which is given as follows. 
% \begin{equation}
%     \mathcal{L} = \mathcal{L}_{CB}(\mathbf{z},y) + \mathcal{L}_{tri}
% \end{equation}
% }

\section{Datasets and Experiments}
\label{sec:datasets}

\subsection{Datasets}
We use three popular benchmarks for the evaluation of LABNet. 
% Each of these benchmarks are widely adopted by the computer vision community. 
Some key comparisons among these datasets are shown in Table \ref{tab:dataset}.
In Fig. \ref{fig:data_dist} (a),(b),(c), the dataset distribution is shown for VERI \cite{liu2017provid}, VehicleID \cite{liu2016deep}, and VERI-Wild \cite{lou2019veri} respectively.  In these figures the id imbalance is clearly visible in all of the three benchmarks. 
A short description of the benchmarks is provided below.

\textbf{VeRi776} \cite{liu2017provid} is one of the most popular benchmarks for vehicle re-identification. VERI consists of 49,325 samples of 776 classes. Among them, 576 classes with 37,746 samples are training samples and the remaining 200 classes with 11,579 samples are for testing. 
The overall data collection is performed with 20 cameras and six viewpoints. 
In the testing set, 1678 images of the 200 vehicles are randomly selected as the query set. During inference, gallery images from the same camera as the probe image are discarded.

\textbf{VehicleID} \cite{liu2016deep} is another standard benchmark for VRID. The dataset contains vehicles with only front or rear viewpoint. It is a large-scale dataset with 221,763 images of 26,267 vehicles. The training set contains 110,178 images of 13,164 vehicles. The testing set is divided into three subsets. The small test subset contains 800 gallery images and 5693 probe images of 800 vehicles. The medium test subset contains 1600 gallery images and 11,777 probe images of 1600 vehicles. The large test subset contains 2400 gallery images and 17,377 probe images of 2400 vehicles.
% Similar to the dataset split mentioned in the benchmark paper, 
When constructing all the probe and gallery subsets, one image is randomly selected for each vehicle and put into the gallery set. Other images of the same vehicle are considered as the probe images. This procedure is repeated until the desired number of samples are drawn, in a manner similar to the dataset split in the benchmark paper.

\textbf{Veri-Wild} \cite{lou2019veri} is a benchmark that is widely adopted by the community. It is a large scale dataset that has been collected under different operating conditions, such as morning, afternoon, night, rainy weather, foggy weather etc. In this benchmark, there are 40,671 vehicles with 416,314 image patches. The dataset was collected with 174 cameras in unconstrained traffic conditions. 
Further, the overall training set contains 30,671 vehicle ids with 277,797 image patches. The testing set is further divided into three subsets, small, medium, and large. 
The small subset has 3,000 images as probe and 41,816 images as gallery of 3,000 vehicles. The medium subset has 5,000 probe images and 69,389 gallery images of 5,000 vehicles. The large subset has 10,000 probe images and 138,517 gallery images of 10,000 vehicles. The procedure for creating these subsets is to randomly select the vehicles from the test set and randomly select 1 image for the probe image from each of these vehicles,
Then the rest of the images of the same vehicle are put in the gallery set.

\begin{table}[htbp]
\caption{Results on VERI dataset} 
\label{tab:veri}
\centering
\setlength\tabcolsep{1.0pt}
{\begin{tabular}{lccr} 
\toprule
Method & mAP & Rank-1 & Venue \\
\midrule
LABNet & \textcolor{red}{84.6} & \textcolor{red}{97.9} & Proposed \\
SPAN+CPDM\cite{chen2020orientation} & 68.9 & 94.0 & ECCV 2020 \\
SAVER\cite{khorramshahi2020devil} & \textcolor{blue}{79.6} & \textcolor{blue}{96.4} & ECCV 2020 \\
PVEN\cite{meng2020parsing} & 79.5 & 95.6 & CVPR 2020 \\
HPGN\cite{shen2020exploring} & \textcolor{cyan}{80.18} & \textcolor{cyan}{96.72} & arXiv 2020 \\
QD-DLF\cite{zhu2019vehicle} & 61.83 & 88.50 & IEEE ITS 2020 \\
\midrule
Appearance+License\cite{he2019combination} & 78.08 & 95.41 & ICIP 2019 \\
SFF+SAtt\cite{liu2019urban} & 74.11 & 94.93 & IGCNN 2019 \\
Part Regularization\cite{he2019part} & 74.30 & 94.30 & CVPR 2019 \\
SAN\cite{qian2020stripe} & 72.50 & 93.30 & arXiv 2019 \\
PAMTRI\cite{tang2019pamtri} & 71.88 & 92.86 & ICCV 2019 \\
MLFN + Triplet\cite{shankar2019comparative} & 71.78 & 92.55 & CVPRW 2019 \\
MTML+Re-ranking\cite{kanaci2019multi} & 68.30 & 90.00 & CVPRW 2019 \\
MRM\cite{peng2019learning} & 68.55 & 91.77 & Neurocomputing 2019\\
DMML\cite{chen2019deep} & 70.10 & 91.20 & ICCV 2019 \\
Triplet Embedding\cite{kuma2019vehicle} & 67.55 & 90.23 & IJCNN 2019 \\
MOV1+BS\cite{tang2019cityflow} & 67.60 & 90.20 & CVPR 2019 \\
VANet\cite{chu2019vehicle} & 66.34 & 89.78 & ICCV 2019 \\
GRF+GGL\cite{liu2019group} & 61.70 & 89.40 & IEEE TIP 2019 \\
ResNet101-AAVER\cite{khorramshahi2019dual} & 61.18 & 88.97 & ICCV 2019 \\
MRM\cite{lin2019multi} & 71.40 & 87.70 & ICME 2019 \\
Fusion-Net\cite{kan2019supervised} & 62.40 & 87.31 & IEEE TIP 2019 \\
Mob. VFL-LSTM\cite{alfasly2019variational} & 58.08 & 87.18 & ICIP 2019 \\
MGL\cite{yang2019vehicle} & 65.00 & 86.10 & ICIP 2019 \\
EALN\cite{lou2019embedding} & 57.44 & 84.39 & IEEE TIP 2019 \\
FDA-Net\cite{lou2019veri} & N/A & 49.43 & CVPR 2019 \\
JDRN+Re-ranking\cite{liu2019supervised} & 73.10 & N/A & CVPRW 2019\\
\midrule
MAD+STR\cite{jiang2018multi} & 61.11 & 89.27 & ICIP 2018 \\
RAM\cite{liu2018ram} & 61.50 & 88.60 & ICME 2018 \\
VAMI\cite{chu2019vehicle} & 61.32 & 88.92 & CVPR 2018 \\
GSTE\cite{bai2018group} & 59.47 & N/A & IEEE TMM \\
SDC-CNN\cite{zhu2018shortly} & 53.45 & 83.49 & ICPR 2018 \\
PROVID\cite{liu2017provid} & 53.42 & 81.56 & IEEE TMM 2018 \\
NuFACT+Plate-SNN\cite{liu2017provid} & 50.87 & 81.11 & IEEE TMM 2018 \\
SCCN-Ft+CLBL-8-Ft\cite{zhou2018vehicle} & 25.12 & 60.83 & IEEE TIP 2018 \\
ABLN-Ft-16\cite{zhou2018v} & 24.92 & 60.49 & WACV 2018 \\
NuFACT\cite{liu2017provid} & 48.47 & 76.76 & IEEE TMM 2018 \\
\midrule
VST Path Proposals\cite{shen2017learning} & 58.27 & 83.49 & ICCV 2017 \\
OIFE+ST\cite{wang2017orientation} & 51.42 & 68.30 & ICCV 2017 \\
DenseNet121\cite{huang2017densely} & 45.06 & 80.27 & CVPR 2017 \\
\midrule
FACT\cite{liu2016deep} & 18.75 & 52.21 & ICME 2016 \\
VGG-CNN-M-1024\cite{liu2016large} & 12.76 & 44.10 & CVPR 2016 \\
GoogLeNet\cite{yang2015large} & 17.89 & 52.32 & CVPR 2016 \\
\bottomrule
\end{tabular}}
%\end{center}
\end{table}

\begin{table*}[hbtp]
\caption{Results on VehicleID dataset} 
\label{tab:vehicleid}
\centering
% \setlength\tabcolsep{0.1pt}
% \resizebox{\textwidth}{!}{\begin{tabular}{lccccccr}
{\begin{tabular}{lccccccr}
\toprule
\multirow{2}{*}{Methods} & \multicolumn{2}{c}{small} &  \multicolumn{2}{c}{medium} & \multicolumn{2}{c}{large} & \multirow{2}{*}{References} \\
& mAP & Rank-1 & mAP & Rank-1 & mAP & Rank-1 & \\
\midrule
LABNet & \textcolor{red}{89.64} & \textcolor{cyan}{84.02} & \textcolor{red}{86.24} & \textcolor{cyan}{80.18} & \textcolor{cyan}{83.48} & \textcolor{blue}{77.2} & Proposed \\
SAVER\cite{khorramshahi2020devil} & N/A & 79.9 & N/A & 77.6 & N/A & 75.3 & ECCV 2020 \\
PVEN\cite{meng2020parsing} &  N/A & \textcolor{red}{84.7} & N/A & \textcolor{red}{80.6} & N/A & \textcolor{red}{77.8} & CVPR 2020 \\ 
HPGN\cite{shen2020exploring} & \textcolor{cyan}{89.60} & \textcolor{blue}{83.91} & \textcolor{cyan}{86.16} & \textcolor{blue}{79.97} & \textcolor{red}{83.60} & \textcolor{cyan}{77.32} & arXiv 2020 \\
QD-DLF\cite{zhu2019vehicle} & 76.54 & 72.32 & 74.63 & 70.66 & 68.41 & 64.14 & IEEE ITS 2020 \\
\midrule
Appearance+License\cite{he2019combination} & 82.7 & 79.5 & 79.9 & 76.9 & 77.7 & 74.8 & ICIP 2019 \\
MGL\cite{yang2019vehicle} & 82.1 & 79.6 & 79.6 & 76.2 & 75.5 & 73.0 & ICIP 2019 \\
Part Regularization\cite{he2019part} & N/A & 78.40 & N/A & 70 & N/A & 74.20 & CVPR 2019 \\
PRN\cite{chen2019partition} & N/A & 78.92 & N/A & 74.94 & N/A & 71.58 & CVPRW 2019 \\
Triplet Embedding\cite{kuma2019vehicle} & \textcolor{blue}{86.19} & 78.80 & \textcolor{blue}{81.69} & 73.41 & \textcolor{blue}{78.16} & 69.33 & IJCNN 2019 \\
MRM\cite{lin2019multi} & 80.02 & 76.64 & 77.32 & 74.20 & 74.02 & 70.86 & Neurocomputing 2019 \\
XG-6-sub-multi \cite{zhao2019structural} & N/A & 76.1 & N/A & 73.1 & N/A & 71.2 & IEEE ITS 2019 \\
GRF+GGL\cite{liu2019group} & N/A & 77.1 & N/A & 72.7 & N/A & 70 & IEEE TIP 2019 \\
MSV\cite{xu2019multi} & 79.3 & 75.1 & 75.4 & 71.8 & 73.3 & 68.7 & ICASSP 2019 \\
DQAL\cite{hou2019deep} & N/A & 74.74 & N/A & 71.01 & N/A & 68.23 & IEEE TVT \\
EALN\cite{lou2019embedding} & 77.5 & 75.11 & 74.2 & 71.78 & 71.0 & 69.30 & IEEE TIP 2019 \\
Mob.VFL-LSTM\cite{alfasly2019variational} & N/A & 73.37 & N/A & 69.52 & N/A & 67.41 & ICIP 2019 \\
ResNet101-AAVER\cite{khorramshahi2019dual} & N/A & 74.69 & N/A & 68.62 & N/A & 63.54 & ICCV 2019 \\
TAMR\cite{guo2019two} & N/A & 66.02 & N/A & 62.90 & N/A & 59.69 & IEEE TIP 2019 \\
MLSR\cite{hou2019multi} & N/A & 65.78 & N/A & 64.24 & N/A & 60.05 & Neurocomputing 2019 \\
RPM\cite{ma2019vehicle} & N/A & 65.04 & N/A & 62.55 & N/A & 60.21 & ICMEW 2019 \\
SFF+SAtt\cite{liu2019urban} & N/A & 64.50 & N/A & 59.12 & N/A & 54.41 & IJCNN 2019 \\
FDA-Net\cite{lou2019veri} & N/A & N/A & 65.33 & 59.84 & 61.84 & 55.53 & CVPR 2019 \\
\midrule
GSTE\cite{bai2018group} & 75.40 & 75.90 & 74.30 & 74.80 & 72.40 & 74.00 & IEEE TMM 2018 \\
RAM\cite{liu2018ram} & N/A & 75.20 & N/A & 72.3 & N/A & 67.70 & ICME 2018 \\
C2F\cite{guo2018learning} & 63.50 & 61.10 & 60.00 & 56.20 & 53 & 51.40 & AAAI 2018 \\
VAMI\cite{chu2019vehicle} & N/A & 63.12 & N/A & 52.87 & N/A & 47.34 & CVPR 2018 \\
SDC-CNN\cite{zhu2018shortly} & 63.52 & 56.98 & 57.07 & 50.57 & 49.68 & 42.92 & ICPR 2018 \\
NuFACT\cite{liu2017provid} & N/A & 48.90 & N/A & 43.64 & N/A & 38.63 & IEEE TMM 2018 \\
MAD+STR\cite{jiang2018multi} & 82.20 & N/A & 75.90 & N/A & 72.80 & N/A & ICIP 2018 \\
PMSM\cite{sun2018part} & 64.20 & N/A & 57.20 & N/A & 51.80 & N/A & ICPR 2018 \\
MSVF\cite{kanaci2018vehicle} & N/A & N/A & N/A & N/A & N/A & 46.61 & GCPR 2018 \\
ABLN-32\cite{zhou2018ve} & N/A & 52.63 & N/A & N/A & N/A & N/A & WACV 2018 \\
\midrule
DJDL\cite{li2017deep} & N/A & 72.30 & N/A & 70.80 & N/A & 68.00 & ICIP 2017 \\
Improved Triplet\cite{zhang2017improving} & N/A & 69.90 & N/A & 66.20 & N/A & 63.20 & ICME 2017 \\
DenseNet121\cite{huang2017densely} & 68.85 & 66.10 & 69.45 & 67.39 & 65.37 & 63.07 & CVPR 2017 \\
MGR\cite{yan2017exploiting} & 62.80 & N/A & 62.30 & N/A & 61.23 & N/A & ICCV 2017 \\
OIFE+ST\cite{wang2017orientation} & N/A & N/A & N/A & N/A & N/A & 67.00 & ICCV 2017 \\
\midrule
DRDL\cite{liu2016deep} & N/A & 48.91 & N/A & 46.36 & N/A & 40.97 & CVPR 2016 \\
FACT\cite{liu2016deep} & N/A & 49.53 & N/A & 44.63 & N/A & 39.91 & ICME 2016 \\
\bottomrule
\end{tabular}}
%\end{center}
\end{table*}

\begin{table*}[hbtp]
\caption{Results on VERI-Wild dataset} 
\label{tab:veriwild}
\centering
% \setlength\tabcolsep{0.2pt}
% \resizebox{0.8\textwidth}{!}{\begin{tabular}{lccccccr}
{\begin{tabular}{lccccccr}
\toprule
\multirow{2}{*}{Methods} & \multicolumn{2}{c}{small} &  \multicolumn{2}{c}{medium} & \multicolumn{2}{c}{large} & \multirow{2}{*}{References} \\
& mAP & Rank-1 & mAP & Rank-1 & mAP & Rank-1 & \\
\midrule
LABNet & \textcolor{red}{82.6} & \textcolor{red}{97.2} & \textcolor{red}{77.4} & \textcolor{red}{96.3} & \textcolor{red}{70.0} & \textcolor{red}{94.5} & Proposed \\
SAVER\cite{khorramshahi2020devil} & \textcolor{blue}{80.9} & \textcolor{blue}{94.5} & \textcolor{blue}{75.3} & \textcolor{blue}{92.7} & \textcolor{blue}{67.7} & \textcolor{blue}{89.5} & ECCV 2020 \\
PVEN\cite{meng2020parsing} & \textcolor{cyan}{82.5} & \textcolor{cyan}{96.7} & \textcolor{cyan}{77.0} & \textcolor{cyan}{95.4} & \textcolor{cyan}{69.7} & \textcolor{cyan}{93.4} & CVPR 2020 \\
HPGN\cite{shen2020exploring} & 80.42 & 91.37 & 75.17 & 88.21 & 65.04 & 82.68 & arXiv 2020\\
Triplet Embedding\cite{kuma2019vehicle} & 70.54 & 84.17 & 62.83 & 78.22 & 51.63 & 69.99 & IJCNN 2019 \\
FDA-Net\cite{lou2019veri} & 35.11 & 64.03 & 29.80 & 57.82 & 22.80 & 49.43 & CVPR 2019 \\
GSTE\cite{bai2018group} & 31.42 & 60.46 & 26.18 & 52.12 & 19.50 & 45.36 & IEEE TMM 2018 \\
Unlabelled GAN\cite{zhu2017unpaired} & 29.86 & 58.06 & 24.71 & 51.58 & 18.23 & 43.63 & ICCV 2017 \\
GoogleNet\cite{yang2015large} & 24.27 & 57.16 & 24.15 & 53.16 & 21.53 & 44.61 & CVPR 2015 \\
HDC\cite{yuan2017hard} & 29.14 & 57.10 & 24.76 & 49.64 & 18.30 & 43.97 & ICCV 2017 \\
DRDL\cite{liu2016deep} & 22.50 & 56.96 & 19.28 & 51.92 & 14.81 & 44.60 & CVPR 2016 \\
Softmax\cite{liu2016de} & 26.41 & 53.40 & 22.66 & 46.16 & 17.62 & 37.94 & ECCV 2016 \\
Triplet\cite{schroff2015facenet} & 15.69 & 44.67 & 13.34 & 14.34 & 9.93 & 33.46 & CVPR 2015 \\

\bottomrule
\end{tabular}}
%\end{center}
\end{table*}

\subsection{Experiments}
We used PyTorch as our implementation toolbox and PyTorch Geometric \cite{FeyLenssen2019} for our Graph module implementation. In our experiments, the resnet101-ibn-a \cite{pan2018IBNNet} was the feature extraction backbone, as it provided superior performance.
We adapted some of our experimental settings from \cite{he2020multi}.
The network was trained with a batch size of 48 and $randomidentity$ sampler during training. The sampler was responsible to randomly select 8 vehicles in each minibatch with 6 samples for each of them. The input image patches were re-scaled to $320\times320$ during both training and testing. For augmentation, the probabilities for both random horizontal flip and random erasing were set to 0.5. The output feature map shape of the backbone is $20\times20$ and our local graph network is built on top of that.
To train our network with VERI and VERI-Wild benchmarks, we started the learning rate at $10^{-3}$ and linearly increased it to $10^{-2}$ in epoch 10. From Epoch 10 to Epoch 39 we kept the learning rate fixed at $10^-{2}$ and reduced the learning rate to $10^{-3}$ at epoch 40 and kept it fixed until epoch 69. At Epoch 70, we reduced the learning rate to $10^{-4}$ and kept it fixed until epoch 120. For the VehicleID dataset, we used 0.1 times the learning rate used for VERI or VERI-WILD dataset. 

We used 0.97 as the class-balanced loss hyperparameter for VERI and VERI-Wild dataset and 0.4 for the VehilceID dataset.
We set the value of $r<2$ in all of our experiments.
The number of Local Graph modules was set to 2 based on the experiments on all the datasets. 
For VehicleID results, we evaluated the performance 5 times and showed the average results for both the mAP and Rank-1 metrics.
Finally, during inference, after accumulating all the query and the gallery images, we utilized the cosine distance for computing the evaluation metrics. 

\section{Evaluation metrics}
\label{sec:evaluation_metrics}
We utilize standard metrics to evaluate the proposed method,
% . The standard performance metrics are 
specifically the mean Average Precision (mAP) and the Cumulative Match Characteristics (CMC).

\textbf{Mean Average Precision} is the primary 
metric for comparing various re-identification algorithms which can be written as follows \cite{lou2019veri}.
\begin{equation}
    AP = \frac{\sum_{r_n=1}^n PR(r_n) \times gt(r_n)}{N_{gt}}
\end{equation}
where, $r_n$ is the rank of a query in a recall list of size $n$, $PR(r_n)$ is the precision within the $r_n$ images and $gt(r_n)$ is the binary value, i.e. 1 if the $r_{n}^{th}$ image is correct else 0. $N_{gt}$ is the number of vehicles with same id as the query image. The mean Average Precision is the mean of all $AP$ for all the query images.

\textbf{Cumulative Matching Characterstics (CMC)} indicates if the top $k_n$ predictions by the network for a query image appear within a gallery list. The CMC at rank $\alpha$ can be written as follows \cite{lou2019veri}.
\begin{equation}
    CMC@\alpha = \frac{\sum_{q=1}^Q gt(q, \alpha)}{Q}
\end{equation}
where $Q$ is the total number of query images and $gt(q,\alpha)$ is 1 if the corresponding query image appears within the rank $\alpha$.

\section{Results}
\label{results}
In this section we compare our method with the state-of-the-art methods and demonstrate quantitative results and the qualitative performance of our method.

\subsection{Quantitative results}
We compare our results with many state-of-the-art methods on the aforementioned benchmarks. We used part of the tables from \cite{shen2020exploring} for comparing with existing methods. 

Table \ref{tab:veri} illustrates the remarkable improvement of VRID techniques on the VERI dataset \cite{liu2017provid} over the years 2016-2019. Compared to the best method of 2016, performance improved by 39.52\% in 2017, 42.75\% in 2018, 59.33\% in 2019.
Our method outperforms the state-of-the-art methods on the VERI dataset. It achieves 4.6\% gain from the HPGN \cite{shen2020exploring} method on mAP metric and about 1.2\% gain on the Rank-1 rate. LABNet also outperforms SAVER \cite{khorramshahi2020devil} and PVEN \cite{meng2020parsing} by more than 5\% on the mAP metric. 

Table \ref{tab:vehicleid} shows the state-of-the-art comparison in the VehicleID \cite{liu2016deep} dataset. We also show the chronological improvement on VRID performance over 2016-2019. LABNet achieves competitive performance in comparison to state-of-the-art methods.
PVEN \cite{meng2020parsing} and HPGN \cite{shen2020exploring} achieved better performance than LABNet by 1.2\% and 0.41\% respectively on Rank-1 metrics. However, LABNet achieves 3.6\% better performance than SAVER \cite{khorramshahi2020devil} and 4\% better performance than Appearance+License \cite{he2019combination} in Rank-1 metrics.

Table \ref{tab:veriwild} presents the VRID performance on the VERI-Wild \cite{lou2019veri} dataset. Our method achieves state-of-the-art results compared to the existing VRID techniques. LABNet outperforms PVEN \cite{meng2020parsing} by 0.1\% and 0.5\% in the small subset on the mAP and Rank-1 metrics. For the medium test subset, LABNet achieved 1.4\% and 0.9\% gain over PVEN \cite{meng2020parsing}. LABNet also outperforms PVEN \cite{meng2020parsing} on the large test subset by 0.3\% map and 1.1\% Rank-1 metrics. The results show that LABNet
% method is capable of 
is performing well on a large scale dataset. Since the VERI-Wild dataset is captured with unconstrained pose of the vehicles, our results indicate that LABNet achieves state-of-the-art pose invariant feature generation.   

\begin{table}[hbtp]
\caption{Ablation study on VERI dataset} 
\label{tab:ablation}
\centering
{\begin{tabular}{lcc}
\toprule
\multirow{2}{*}{Methods} & \multicolumn{2}{c}{VERI} \\
& mAP & Rank-1 \\
\midrule
Baseline+RE+BN+LGA+CB & 84.6 & 97.9  \\
Baseline+RE+BN+LGA & 84.6 & 97.3  \\
Baseline+RE+BN+LAP & 83.1 & 96.5  \\
Baseline+RE+BN & 83.5 & 97.2  \\
Baseline+RE & 78.9 & 96.1 \\
Baseline & 75.2 & 94.2 \\
\bottomrule
\end{tabular}}
%\end{center}
\end{table}

Table \ref{tab:ablation} shows how each part of our proposed network improves performance on the VERI \cite{liu2017provid} dataset. First,  the baseline itself achieved superior performance due to the backbone, data augmentation techniques, and training procedure. The addition of augmentation with Random Erasing (RE) further improved the mAP by 3.7\%. The incorporation of the BN neck improved the mAP by 3.6\%. We also evaluated the feature aggregation performance using Local Average Pooling (LAP) instead of graph based aggregation to compare the performance. However, LAP reduced performance by 0.4\% over the baseline on mAP. The Rank-1 performance was reduced by 0.7\%.
We then utilized our LGA module in the network and it improved performance by 1.1\% on mAP. With the addition of Class Balanced (CB) loss, we achieved 0.6\% improvement on Rank-1 performance compared to the Baseline+RE+BN+LGA model. Overall, our model improved the performance by 11.4\% on mAP and 5.7\% on Rank-1 metric compared to the Baseline model. These experiments demonstrate the effectiveness of our method. 

\begin{table}[hbtp]
\caption{Varying LGA on VehicleID dataset} 
\label{tab:vehicleidablation}
\centering
\resizebox{0.48\textwidth}{!}{\begin{tabular}{ccccccr}
% {\begin{tabular}{ccccccr}
\toprule
\multirow{2}{*}{LGA Modules} & \multicolumn{2}{c}{small} &  \multicolumn{2}{c}{medium} & \multicolumn{2}{c}{large} \\
& mAP & Rank-1 & mAP & Rank-1 & mAP & Rank-1 \\
\midrule
1 & 89.6 & 83.9 & 86.36 & 80.52 & 83.76 & 77.44 \\
1+2 & 89.64 & 84.02 & 86.24 & 80.18 & 83.48 & 77.2 \\
1+2+3 & 89.46 & 84.0 & 86.08 & 80.06 & 83.84 & 77.56 \\
1+2+3+4 & 89.94 & 84.56 & 86.5 & 80.7 & 83.5 & 77.14 \\
1+2+3+4+5 & 90.2 & 85.04 & 85.86 & 79.62 & 83.54 & 77.2 \\
\bottomrule
\end{tabular}}
%\end{center}
\end{table}

\begin{figure*}[hbtp]
\centerline{\includegraphics[width=0.8\textwidth]{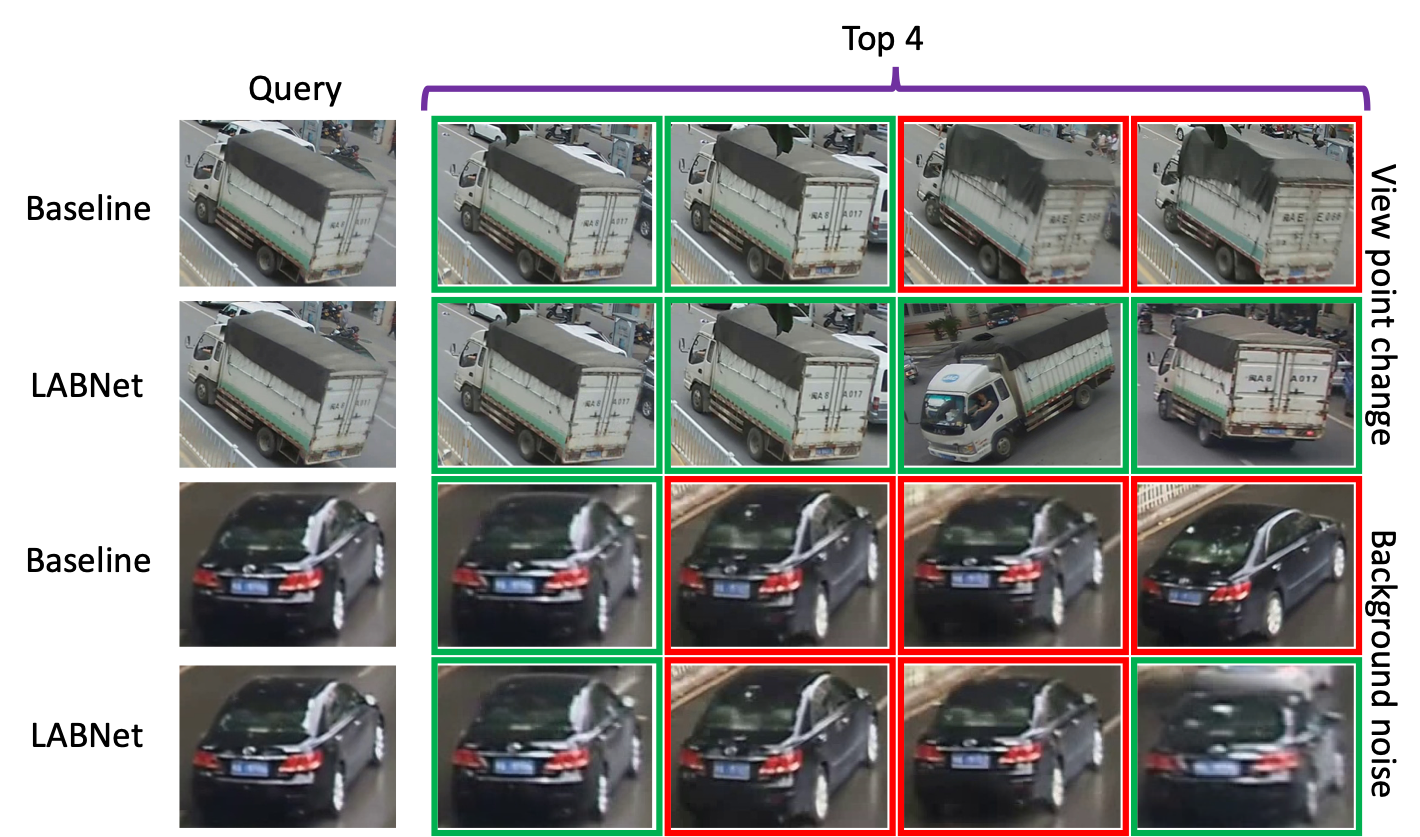}}
\caption{Visual results. Red represents wrong matching and blue represents correct matching. Best viewed in color.}
\label{fig:samples}
\end{figure*}

% \begin{figure}[hbtp]
% \centerline{\includegraphics[width=0.3\textwidth]{images/grad_cam.png}}
% \caption{Activation map visualization for the baseline and LABNet.}
% \label{fig:gradcam}
% \end{figure}

% \begin{figure}[hbtp]
% \centerline{\includegraphics[width=0.4\textwidth]{images/tsne.png}}
% \caption{Visualization of the learned feature space of LABNet.}
% \label{fig:tsne}
% \end{figure}

Table \ref{tab:vehicleidablation} shows the results of further experiments that evaluate the performance for various depths of graph modules on the VehicleID \cite{liu2016deep} datasets. We found in our experiments that cascading multiple graph modules is beneficial on the small test set. However, the performance on the large test subsets is not affected by
% provides similar performance irrespective of 
the number of LGA modules.

\subsection{Qualitative results and visualization}

Fig. \ref{fig:samples} shows two successful cases of the proposed LABNet algorithm, where the top 4 predictions are included with the corresponding query images.
% in Figure \ref{fig:samples}. 
For the partial occlusion scenario, LABNet successfully identifies the correct vehicle when the query image or the target images are partially occluded.  
For the viewpoint change scenario, the baseline method got confused with a vehicle having similar appearance and similar viewpoint. LABNet was able to successfully identify the correct vehicle even though there is slight viewpoint change in the third and fourth position.

\section{Conclusion}
\label{sec:conclusion}
% Vehicle re-identification has significant applications in video surveillance and smart traffic control. 
In recent years, deep learning methods achieved significant improvements in the field of vehicle re-identification on various large scale benchmarks. 
In this paper, we proposed LABNet, a deep learning based method that further advances vehicle re-identification. Our method utilizes graph based feature aggregation on the backbone feature map to aggregate information explicitly among various spatial locations. 
We incorporate a class balanced loss with the softmax loss during training that improved feature learning on imbalanced datasets. We performed various experiments on multiple benchmarks and our method generally outperforms state-of-the-art methods. 
Finally, we show qualitative results to demonstrate the effectiveness of the proposed method.     

\section*{Acknowledgments}
This research was supported in part by the Air Force Research Laboratory, Sensors Directorate (AFRL/RYAP) under contract number FA8650-18-C-1739 to Systems and Technology Research. 
The authors acknowledge the computational resources made available by Research Computing at Rochester Institute of Technology that helped produce part of the results.

{\small
\bibliographystyle{ieee.bst}
\bibliography{./refs}

\begin{thebibliography}{10}\itemsep=-1pt

\bibitem{alfasly2019variational}
S.~A.~S. Alfasly, Y.~Hu, T.~Liang, X.~Jin, Q.~Zhao, and B.~Liu.
\newblock Variational representation learning for vehicle re-identificati.
\newblock In {\em 2019 IEEE International Conference on Image Processing
  (ICIP)}, pages 3118--3122. IEEE, 2019.

\bibitem{anagnostopoulos2008license}
C.-N.~E. Anagnostopoulos, I.~E. Anagnostopoulos, I.~D. Psoroulas, V.~Loumos,
  and E.~Kayafas.
\newblock License plate recognition from still images and video sequences: A
  survey.
\newblock {\em IEEE Transactions on intelligent transportation systems},
  9(3):377--391, 2008.

\bibitem{bai2018group}
Y.~Bai, Y.~Lou, F.~Gao, S.~Wang, Y.~Wu, and L.-Y. Duan.
\newblock Group-sensitive triplet embedding for vehicle reidentification.
\newblock {\em IEEE Transactions on Multimedia}, 20(9):2385--2399, 2018.

\bibitem{bao2019masked}
L.~Bao, B.~Ma, H.~Chang, and X.~Chen.
\newblock Masked graph attention network for person re-identification.
\newblock In {\em Proceedings of the IEEE Conference on Computer Vision and
  Pattern Recognition Workshops}, pages 0--0, 2019.

\bibitem{chen2019deep}
G.~Chen, T.~Zhang, J.~Lu, and J.~Zhou.
\newblock Deep meta metric learning.
\newblock In {\em Proceedings of the IEEE International Conference on Computer
  Vision}, pages 9547--9556, 2019.

\bibitem{chen2019partition}
H.~Chen, B.~Lagadec, and F.~Bremond.
\newblock Partition and reunion: A two-branch neural network for vehicle
  re-identification.
\newblock In {\em CVPR Workshops}, pages 184--192, 2019.

\bibitem{chen2020orientation}
T.-S. Chen, C.-T. Liu, C.-W. Wu, and S.-Y. Chien.
\newblock Orientation-aware vehicle re-identification with semantics-guided
  part attention network.
\newblock {\em arXiv preprint arXiv:2008.11423}, 2020.

\bibitem{chu2019vehicle}
R.~Chu, Y.~Sun, Y.~Li, Z.~Liu, C.~Zhang, and Y.~Wei.
\newblock Vehicle re-identification with viewpoint-aware metric learning.
\newblock In {\em Proceedings of the IEEE International Conference on Computer
  Vision}, pages 8282--8291, 2019.

\bibitem{cui2019class}
Y.~Cui, M.~Jia, T.-Y. Lin, Y.~Song, and S.~Belongie.
\newblock Class-balanced loss based on effective number of samples.
\newblock In {\em Proceedings of the IEEE Conference on Computer Vision and
  Pattern Recognition}, pages 9268--9277, 2019.

\bibitem{FeyLenssen2019}
M.~Fey and J.~E. Lenssen.
\newblock Fast graph representation learning with {PyTorch Geometric}.
\newblock In {\em ICLR Workshop on Representation Learning on Graphs and
  Manifolds}, 2019.

\bibitem{guo2018learning}
H.~Guo, C.~Zhao, Z.~Liu, J.~Wang, and H.~Lu.
\newblock Learning coarse-to-fine structured feature embedding for vehicle
  re-identification.
\newblock In {\em Thirty-Second AAAI Conference on Artificial Intelligence},
  2018.

\bibitem{guo2019two}
H.~Guo, K.~Zhu, M.~Tang, and J.~Wang.
\newblock Two-level attention network with multi-grain ranking loss for vehicle
  re-identification.
\newblock {\em IEEE Transactions on Image Processing}, 28(9):4328--4338, 2019.

\bibitem{he2019part}
B.~He, J.~Li, Y.~Zhao, and Y.~Tian.
\newblock Part-regularized near-duplicate vehicle re-identification.
\newblock In {\em Proceedings of the IEEE Conference on Computer Vision and
  Pattern Recognition}, pages 3997--4005, 2019.

\bibitem{he2020multi}
S.~He, H.~Luo, W.~Chen, M.~Zhang, Y.~Zhang, F.~Wang, H.~Li, and W.~Jiang.
\newblock Multi-domain learning and identity mining for vehicle
  re-identification.
\newblock In {\em Proceedings of the IEEE/CVF Conference on Computer Vision and
  Pattern Recognition Workshops}, pages 582--583, 2020.

\bibitem{he2019combination}
Y.~He, C.~Dong, and Y.~Wei.
\newblock Combination of appearance and license plate features for vehicle
  re-identification.
\newblock In {\em 2019 IEEE International Conference on Image Processing
  (ICIP)}, pages 3108--3112. IEEE, 2019.

\bibitem{hermans2017defense}
A.~Hermans, L.~Beyer, and B.~Leibe.
\newblock In defense of the triplet loss for person re-identification.
\newblock {\em arXiv preprint arXiv:1703.07737}, 2017.

\bibitem{hou2019multi}
J.~Hou, H.~Zeng, L.~Cai, J.~Zhu, J.~Chen, and K.-K. Ma.
\newblock Multi-label learning with multi-label smoothing regularization for
  vehicle re-identification.
\newblock {\em Neurocomputing}, 345:15--22, 2019.

\bibitem{hou2019deep}
J.~Hou, H.~Zeng, J.~Zhu, J.~Hou, J.~Chen, and K.-K. Ma.
\newblock Deep quadruplet appearance learning for vehicle re-identification.
\newblock {\em IEEE Transactions on Vehicular Technology}, 68(9):8512--8522,
  2019.

\bibitem{huang2017densely}
G.~Huang, Z.~Liu, L.~Van Der~Maaten, and K.~Q. Weinberger.
\newblock Densely connected convolutional networks.
\newblock In {\em Proceedings of the IEEE conference on computer vision and
  pattern recognition}, pages 4700--4708, 2017.

\bibitem{jiang2018multi}
N.~Jiang, Y.~Xu, Z.~Zhou, and W.~Wu.
\newblock Multi-attribute driven vehicle re-identification with
  spatial-temporal re-ranking.
\newblock In {\em 2018 25th IEEE International Conference on Image Processing
  (ICIP)}, pages 858--862. IEEE, 2018.

\bibitem{johnson2019survey}
J.~M. Johnson and T.~M. Khoshgoftaar.
\newblock Survey on deep learning with class imbalance.
\newblock {\em Journal of Big Data}, 6(1):27, 2019.

\bibitem{kan2019supervised}
S.~Kan, Y.~Cen, Z.~He, Z.~Zhang, L.~Zhang, and Y.~Wang.
\newblock Supervised deep feature embedding with handcrafted feature.
\newblock {\em IEEE Transactions on Image Processing}, 28(12):5809--5823, 2019.

\bibitem{kanaci2019multi}
A.~Kanaci, M.~Li, S.~Gong, and G.~Rajamanoharan.
\newblock Multi-task mutual learning for vehicle re-identification.
\newblock In {\em Proceedings of the IEEE Conference on Computer Vision and
  Pattern Recognition Workshops}, pages 62--70, 2019.

\bibitem{kanaci2018vehicle}
A.~Kanac{\i}, X.~Zhu, and S.~Gong.
\newblock Vehicle re-identification in context.
\newblock In {\em German Conference on Pattern Recognition}, pages 377--390.
  Springer, 2018.

\bibitem{khan2019survey}
S.~D. Khan and H.~Ullah.
\newblock A survey of advances in vision-based vehicle re-identification.
\newblock {\em Computer Vision and Image Understanding}, 182:50--63, 2019.

\bibitem{khorramshahi2019dual}
P.~Khorramshahi, A.~Kumar, N.~Peri, S.~S. Rambhatla, J.-C. Chen, and
  R.~Chellappa.
\newblock A dual-path model with adaptive attention for vehicle
  re-identification.
\newblock In {\em Proceedings of the IEEE International Conference on Computer
  Vision}, pages 6132--6141, 2019.

\bibitem{khorramshahi2020devil}
P.~Khorramshahi, N.~Peri, J.-c. Chen, and R.~Chellappa.
\newblock The devil is in the details: Self-supervised attention for vehicle
  re-identification.
\newblock {\em arXiv preprint arXiv:2004.06271}, 2020.

\bibitem{kipf2016semi}
T.~N. Kipf and M.~Welling.
\newblock Semi-supervised classification with graph convolutional networks.
\newblock {\em arXiv preprint arXiv:1609.02907}, 2016.

\bibitem{kuma2019vehicle}
R.~Kuma, E.~Weill, F.~Aghdasi, and P.~Sriram.
\newblock Vehicle re-identification: an efficient baseline using triplet
  embedding.
\newblock In {\em 2019 International Joint Conference on Neural Networks
  (IJCNN)}, pages 1--9. IEEE, 2019.

\bibitem{li2017deep}
Y.~Li, Y.~Li, H.~Yan, and J.~Liu.
\newblock Deep joint discriminative learning for vehicle re-identification and
  retrieval.
\newblock In {\em 2017 IEEE International Conference on Image Processing
  (ICIP)}, pages 395--399. IEEE, 2017.

\bibitem{lin2019multi}
W.~Lin, Y.~Li, X.~Yang, P.~Peng, and J.~Xing.
\newblock Multi-view learning for vehicle re-identification.
\newblock In {\em 2019 IEEE International Conference on Multimedia and Expo
  (ICME)}, pages 832--837. IEEE, 2019.

\bibitem{liu2019urban}
C.~Liu, D.~Q. Huynh, and M.~Reynolds.
\newblock Urban area vehicle re-identification with self-attention stair
  feature fusion and temporal bayesian re-ranking.
\newblock In {\em 2019 International Joint Conference on Neural Networks
  (IJCNN)}, pages 1--8. IEEE, 2019.

\bibitem{liu2019supervised}
C.-T. Liu, M.-Y. Lee, C.-W. Wu, B.-Y. Chen, T.-S. Chen, Y.-T. Hsu, S.-Y. Chien,
  and N.~I. Center.
\newblock Supervised joint domain learning for vehicle re-identification.
\newblock In {\em CVPR Workshops}, pages 45--52, 2019.

\bibitem{liu2016deep}
H.~Liu, Y.~Tian, Y.~Yang, L.~Pang, and T.~Huang.
\newblock Deep relative distance learning: Tell the difference between similar
  vehicles.
\newblock In {\em Proceedings of the IEEE Conference on Computer Vision and
  Pattern Recognition}, pages 2167--2175, 2016.

\bibitem{liu2016large}
X.~Liu, W.~Liu, H.~Ma, and H.~Fu.
\newblock Large-scale vehicle re-identification in urban surveillance videos.
\newblock In {\em 2016 IEEE International Conference on Multimedia and Expo
  (ICME)}, pages 1--6. IEEE, 2016.

\bibitem{liu2016de}
X.~Liu, W.~Liu, T.~Mei, and H.~Ma.
\newblock A deep learning-based approach to progressive vehicle
  re-identification for urban surveillance.
\newblock In {\em European conference on computer vision}, pages 869--884.
  Springer, 2016.

\bibitem{liu2017provid}
X.~Liu, W.~Liu, T.~Mei, and H.~Ma.
\newblock Provid: Progressive and multimodal vehicle reidentification for
  large-scale urban surveillance.
\newblock {\em IEEE Transactions on Multimedia}, 20(3):645--658, 2017.

\bibitem{liu2018ram}
X.~Liu, S.~Zhang, Q.~Huang, and W.~Gao.
\newblock Ram: a region-aware deep model for vehicle re-identification.
\newblock In {\em 2018 IEEE International Conference on Multimedia and Expo
  (ICME)}, pages 1--6. IEEE, 2018.

\bibitem{liu2019group}
X.~Liu, S.~Zhang, X.~Wang, R.~Hong, and Q.~Tian.
\newblock Group-group loss-based global-regional feature learning for vehicle
  re-identification.
\newblock {\em IEEE Transactions on Image Processing}, 29:2638--2652, 2019.

\bibitem{lou2019veri}
Y.~Lou, Y.~Bai, J.~Liu, S.~Wang, and L.~Duan.
\newblock Veri-wild: A large dataset and a new method for vehicle
  re-identification in the wild.
\newblock In {\em Proceedings of the IEEE Conference on Computer Vision and
  Pattern Recognition}, pages 3235--3243, 2019.

\bibitem{lou2019embedding}
Y.~Lou, Y.~Bai, J.~Liu, S.~Wang, and L.-Y. Duan.
\newblock Embedding adversarial learning for vehicle re-identification.
\newblock {\em IEEE Transactions on Image Processing}, 28(8):3794--3807, 2019.

\bibitem{luo2019strongtmm}
H.~{Luo}, W.~{Jiang}, Y.~{Gu}, F.~{Liu}, X.~{Liao}, S.~{Lai}, and J.~{Gu}.
\newblock A strong baseline and batch normalization neck for deep person
  re-identification.
\newblock {\em IEEE Transactions on Multimedia}, pages 1--1, 2019.

\bibitem{ma2019vehicle}
X.~Ma, K.~Zhu, H.~Guo, J.~Wang, M.~Huang, and Q.~Miao.
\newblock Vehicle re-identification with refined part model.
\newblock In {\em 2019 IEEE International Conference on Multimedia \& Expo
  Workshops (ICMEW)}, pages 603--606. IEEE, 2019.

\bibitem{meng2020parsing}
D.~Meng, L.~Li, X.~Liu, Y.~Li, S.~Yang, Z.-J. Zha, X.~Gao, S.~Wang, and
  Q.~Huang.
\newblock Parsing-based view-aware embedding network for vehicle
  re-identification.
\newblock In {\em Proceedings of the IEEE/CVF Conference on Computer Vision and
  Pattern Recognition}, pages 7103--7112, 2020.

\bibitem{pan2018IBNNet}
X.~Pan, P.~Luo, J.~Shi, and X.~Tang.
\newblock Two at once: Enhancing learning and generalization capacities via
  ibn-net.
\newblock In {\em ECCV}, 2018.

\bibitem{peng2019learning}
J.~Peng, H.~Wang, T.~Zhao, and X.~Fu.
\newblock Learning multi-region features for vehicle re-identification with
  context-based ranking method.
\newblock {\em Neurocomputing}, 359:427--437, 2019.

\bibitem{porrello2020robust}
A.~Porrello, L.~Bergamini, and S.~Calderara.
\newblock Robust re-identification by multiple views knowledge distillation.
\newblock {\em arXiv preprint arXiv:2007.04174}, 2020.

\bibitem{qian2020stripe}
J.~Qian, W.~Jiang, H.~Luo, and H.~Yu.
\newblock Stripe-based and attribute-aware network: A two-branch deep model for
  vehicle re-identification.
\newblock {\em Measurement Science and Technology}, 2020.

\bibitem{quan2019auto}
R.~Quan, X.~Dong, Y.~Wu, L.~Zhu, and Y.~Yang.
\newblock Auto-reid: Searching for a part-aware convnet for person
  re-identification.
\newblock In {\em Proceedings of the IEEE International Conference on Computer
  Vision}, pages 3750--3759, 2019.

\bibitem{schroff2015facenet}
F.~Schroff, D.~Kalenichenko, and J.~Philbin.
\newblock Facenet: A unified embedding for face recognition and clustering.
\newblock In {\em Proceedings of the IEEE conference on computer vision and
  pattern recognition}, pages 815--823, 2015.

\bibitem{shankar2019comparative}
A.~Shankar, A.~Poojary, V.~Kollerathu, C.~Yeshwanth, S.~Reddy, and
  V.~Sudhakaran.
\newblock Comparative study on various losses for vehicle re-identification.
\newblock In {\em CVPR Workshops}, volume~2, 2019.

\bibitem{shen2020exploring}
F.~Shen, J.~Zhu, X.~Zhu, Y.~Xie, and J.~Huang.
\newblock Exploring spatial significance via hybrid pyramidal graph network for
  vehicle re-identification.
\newblock {\em arXiv preprint arXiv:2005.14684}, 2020.

\bibitem{shen2017learning}
Y.~Shen, T.~Xiao, H.~Li, S.~Yi, and X.~Wang.
\newblock Learning deep neural networks for vehicle re-id with
  visual-spatio-temporal path proposals.
\newblock In {\em Proceedings of the IEEE International Conference on Computer
  Vision}, pages 1900--1909, 2017.

\bibitem{sun2018part}
Y.~Sun, M.~Li, and J.~Lu.
\newblock Part-based multi-stream model for vehicle searching.
\newblock In {\em 2018 24th International Conference on Pattern Recognition
  (ICPR)}, pages 1372--1377. IEEE, 2018.

\bibitem{sun2014deep}
Y.~Sun, X.~Wang, and X.~Tang.
\newblock Deep learning face representation from predicting 10,000 classes.
\newblock In {\em Proceedings of the IEEE conference on computer vision and
  pattern recognition}, pages 1891--1898, 2014.

\bibitem{tang2019pamtri}
Z.~Tang, M.~Naphade, S.~Birchfield, J.~Tremblay, W.~Hodge, R.~Kumar, S.~Wang,
  and X.~Yang.
\newblock Pamtri: Pose-aware multi-task learning for vehicle re-identification
  using highly randomized synthetic data.
\newblock In {\em Proceedings of the IEEE International Conference on Computer
  Vision}, pages 211--220, 2019.

\bibitem{tang2019cityflow}
Z.~Tang, M.~Naphade, M.-Y. Liu, X.~Yang, S.~Birchfield, S.~Wang, R.~Kumar,
  D.~Anastasiu, and J.-N. Hwang.
\newblock Cityflow: A city-scale benchmark for multi-target multi-camera
  vehicle tracking and re-identification.
\newblock In {\em Proceedings of the IEEE Conference on Computer Vision and
  Pattern Recognition}, pages 8797--8806, 2019.

\bibitem{taufique2020benchmarking}
A.~M.~N. Taufique, B.~Minnehan, and A.~Savakis.
\newblock Benchmarking deep trackers on aerial videos.
\newblock {\em Sensors}, 20(2):547, 2020.

\bibitem{wang2017orientation}
Z.~Wang, L.~Tang, X.~Liu, Z.~Yao, S.~Yi, J.~Shao, J.~Yan, S.~Wang, H.~Li, and
  X.~Wang.
\newblock Orientation invariant feature embedding and spatial temporal
  regularization for vehicle re-identification.
\newblock In {\em Proceedings of the IEEE International Conference on Computer
  Vision}, pages 379--387, 2017.

\bibitem{watcharapinchai2017approximate}
N.~Watcharapinchai and S.~Rujikietgumjorn.
\newblock Approximate license plate string matching for vehicle
  re-identification.
\newblock In {\em 2017 14th IEEE International Conference on Advanced Video and
  Signal Based Surveillance (AVSS)}, pages 1--6. IEEE, 2017.

\bibitem{xu2019multi}
Y.~Xu, N.~Jiang, L.~Zhang, Z.~Zhou, and W.~Wu.
\newblock Multi-scale vehicle re-identification using self-adapting label
  smoothing regularization.
\newblock In {\em ICASSP 2019-2019 IEEE International Conference on Acoustics,
  Speech and Signal Processing (ICASSP)}, pages 2117--2121. IEEE, 2019.

\bibitem{yan2017exploiting}
K.~Yan, Y.~Tian, Y.~Wang, W.~Zeng, and T.~Huang.
\newblock Exploiting multi-grain ranking constraints for precisely searching
  visually-similar vehicles.
\newblock In {\em Proceedings of the IEEE International Conference on Computer
  Vision}, pages 562--570, 2017.

\bibitem{yang2015large}
L.~Yang, P.~Luo, C.~Change~Loy, and X.~Tang.
\newblock A large-scale car dataset for fine-grained categorization and
  verification.
\newblock In {\em Proceedings of the IEEE conference on computer vision and
  pattern recognition}, pages 3973--3981, 2015.

\bibitem{yang2019vehicle}
X.~Yang, C.~Lang, P.~Peng, and J.~Xing.
\newblock Vehicle re-identification by multi-grain learni.
\newblock In {\em 2019 IEEE International Conference on Image Processing
  (ICIP)}, pages 3113--3117. IEEE, 2019.

\bibitem{yuan2017hard}
Y.~Yuan, K.~Yang, and C.~Zhang.
\newblock Hard-aware deeply cascaded embedding.
\newblock In {\em Proceedings of the IEEE international conference on computer
  vision}, pages 814--823, 2017.

\bibitem{zapletal2016vehicle}
D.~Zapletal and A.~Herout.
\newblock Vehicle re-identification for automatic video traffic surveillance.
\newblock In {\em Proceedings of the IEEE Conference on Computer Vision and
  Pattern Recognition Workshops}, pages 25--31, 2016.

\bibitem{zhang2017improving}
Y.~Zhang, D.~Liu, and Z.-J. Zha.
\newblock Improving triplet-wise training of convolutional neural network for
  vehicle re-identification.
\newblock In {\em 2017 IEEE International Conference on Multimedia and Expo
  (ICME)}, pages 1386--1391. IEEE, 2017.

\bibitem{zhao2019structural}
Y.~Zhao, C.~Shen, H.~Wang, and S.~Chen.
\newblock Structural analysis of attributes for vehicle re-identification and
  retrieval.
\newblock {\em IEEE Transactions on Intelligent Transportation Systems},
  21(2):723--734, 2019.

\bibitem{zheng2020going}
Z.~Zheng, M.~Jiang, Z.~Wang, J.~Wang, Z.~Bai, X.~Zhang, X.~Yu, X.~Tan, Y.~Yang,
  S.~Wen, et~al.
\newblock Going beyond real data: A robust visual representation for vehicle
  re-identification.
\newblock In {\em Proceedings of the IEEE/CVF Conference on Computer Vision and
  Pattern Recognition Workshops}, pages 598--599, 2020.

\bibitem{zheng2020vehiclenet}
Z.~Zheng, T.~Ruan, Y.~Wei, Y.~Yang, and T.~Mei.
\newblock Vehiclenet: Learning robust visual representation for vehicle
  re-identification.
\newblock {\em arXiv preprint arXiv:2004.06305}, 2020.

\bibitem{zhou2018vehicle}
Y.~Zhou, L.~Liu, and L.~Shao.
\newblock Vehicle re-identification by deep hidden multi-view inference.
\newblock {\em IEEE Transactions on Image Processing}, 27(7):3275--3287, 2018.

\bibitem{zhou2018v}
Y.~Zhou and L.~Shao.
\newblock Vehicle re-identification by adversarial bi-directional lstm network.
\newblock In {\em 2018 IEEE Winter Conference on Applications of Computer
  Vision (WACV)}, pages 653--662. IEEE, 2018.

\bibitem{zhou2018ve}
Y.~Zhou and L.~Shao.
\newblock Vehicle re-identification by adversarial bi-directional lstm network.
\newblock In {\em 2018 IEEE Winter Conference on Applications of Computer
  Vision (WACV)}, pages 653--662. IEEE, 2018.

\bibitem{zhu2019vehicle}
J.~Zhu, H.~Zeng, J.~Huang, S.~Liao, Z.~Lei, C.~Cai, and L.~Zheng.
\newblock Vehicle re-identification using quadruple directional deep learning
  features.
\newblock {\em IEEE Transactions on Intelligent Transportation Systems},
  21(1):410--420, 2019.

\bibitem{zhu2018shortly}
J.~Zhu, H.~Zeng, Z.~Lei, S.~Liao, L.~Zheng, and C.~Cai.
\newblock A shortly and densely connected convolutional neural network for
  vehicle re-identification.
\newblock In {\em 2018 24th International Conference on Pattern Recognition
  (ICPR)}, pages 3285--3290. IEEE, 2018.

\bibitem{zhu2017unpaired}
J.-Y. Zhu, T.~Park, P.~Isola, and A.~A. Efros.
\newblock Unpaired image-to-image translation using cycle-consistent
  adversarial networks.
\newblock In {\em Proceedings of the IEEE international conference on computer
  vision}, pages 2223--2232, 2017.

\end{thebibliography}
}

\end{document}